%% file: main.tex
\documentclass[runningheads]{llncs}
\usepackage{hhline}
\usepackage{multirow}
\usepackage{booktabs}
\usepackage{caption}
\usepackage{subcaption}
\usepackage{svg}
\usepackage{amsmath,amsfonts,bm} 

\newcommand{\vect}[1]{\mathbf{#1}}
\usepackage{mathtools,etoolbox}
\usepackage{graphicx}

\begin{document}
\title{Improving Aleatoric Uncertainty Quantification in Multi-Annotated Medical Image Segmentation with Normalizing Flows}
%
%
\author{M.M.A. Valiuddin\inst{1} \and C.G.A. Viviers\inst{1} \and\ R.J.G. van Sloun\inst{1} \and P.H.N. de With\inst{1} \and F. van der Sommen\inst{1}}

\authorrunning{M.M.A. Valiuddin et al.}
%
\institute{$^1$Eindhoven University of Technology, Eindhoven 5612 AZ, The Netherlands \email{m.m.a.valiuddin@student.tue.nl}}
\maketitle              
\begin{abstract}
\vspace{-0.5cm}
Quantifying uncertainty in medical image segmentation applications is essential, as it is often connected to vital decision-making. Compelling attempts have been made in quantifying the uncertainty in image segmentation architectures, e.g. to learn a density segmentation model conditioned on the input image. Typical work in this field restricts these learnt densities to be strictly Gaussian. In this paper, we propose to use a more flexible approach by introducing Normalizing Flows (NFs), which enables the learnt densities to be more complex and facilitate more accurate modeling for uncertainty. We prove this hypothesis by adopting the Probabilistic U-Net and augmenting the posterior density with an NF, allowing it to be more expressive. Our qualitative as well as quantitative (GED and IoU) evaluations on the multi-annotated and single-annotated LIDC-IDRI and Kvasir-SEG segmentation datasets, respectively, show a clear improvement. This is mostly apparent in the quantification of aleatoric uncertainty and the increased predictive performance of up to 14 percent. This result strongly indicates that a more flexible density model should be seriously considered in architectures that attempt to capture segmentation ambiguity through density modeling. The benefit of this improved modeling will increase human confidence in annotation and segmentation, and enable eager adoption of the technology in practice.
\vspace{-0.25cm}
\end{abstract}
\input{introduction}
\input{relatedwork}
\input{architecture}
\input{method}
\input{results}
\input{conclusion}
%
%
%

%
\clearpage
\bibliographystyle{splncs04}
\bibliography{bibliography}
\clearpage
\appendix
\chapter*{Appendices}
\section{Probabilistic U-Net objective}\label{elboloss}
The loss function of the PU-Net is based on the standard ELBO and is defined as
\begin{equation}
    \begin{aligned}
        \mathcal{L} &= -\mathbb{E}_{q_\phi(\mathbf{z}\vert\mathbf{s},\mathbf{x})}[\,\operatorname{log}p(\mathbf{s}\vert\mathbf{z},\mathbf{x})\,]+\operatorname{KL}\left(\,q_\phi(\mathbf{z}\vert\mathbf{s},\mathbf{x})\vert\vert p_\psi(\mathbf{z}\vert\mathbf{x})\,\right),
    \end{aligned}
    \label{elbo}
\end{equation}
where the latent sample $\mathbf{z}$ from the posterior distribution is conditioned on the input image $\mathbf{x}$, and ground-truth segmentation $\mathbf{s}$.
\section{Planar and radial flows}\label{flows}
Normalizing Flows are trained by maximizing the likelihood objective
\begin{equation}
    \begin{aligned}
        \log p(\mathbf{x})=\log p_{0}\left(\mathbf{z}_{0}\right)-\sum_{i=1}^{K} \log \left(\left|\operatorname{det} \frac{d f_{i}}{d \mathbf{z}_{i-1}}\right|\right).
    \end{aligned}
    \label{eq:loglikely}
\end{equation}
In the PU-Net, the objective becomes
\begin{equation}
    \begin{aligned}
        \log q(\mathbf{z}\vert\mathbf{s},\mathbf{x})=\log q_{0}(\mathbf{z}_0\vert\mathbf{s},\mathbf{x})-\sum_{i=1}^{K} \log \left(\left|\operatorname{det} \frac{d f_{i}}{d \mathbf{z}_{i-1}}\right|\right),
    \end{aligned}
\end{equation}
where the \textit{i}-th latent sample $\mathbf{z}_i$ from the Normalizing Flow is conditioned on the input image $\mathbf{x}$, and ground-truth segmentation $\mathbf{s}$.\\

\noindent
The planar flow expands and contracts distributions along a specific directions by applying the transformation
\begin{equation}
        f(\vect{x})=\vect{x} + \vect{u}h(\vect{w}^T\vect{x}+\vect{b}),
        \label{eq:pflow}
\end{equation}
while the radial flow warps distributions around a specific point with the transformation
\begin{equation}
        f(\vect{x})=\vect{x} + \frac{\beta}{\alpha \left|\vect{x}-\vect{x}_0\right|}(\vect{x}-\vect{x}_0).
\end{equation}
\newpage
\section{Dataset images}\label{dataset}
\begin{figure}
  \centerline{\includegraphics[width=0.7\linewidth]{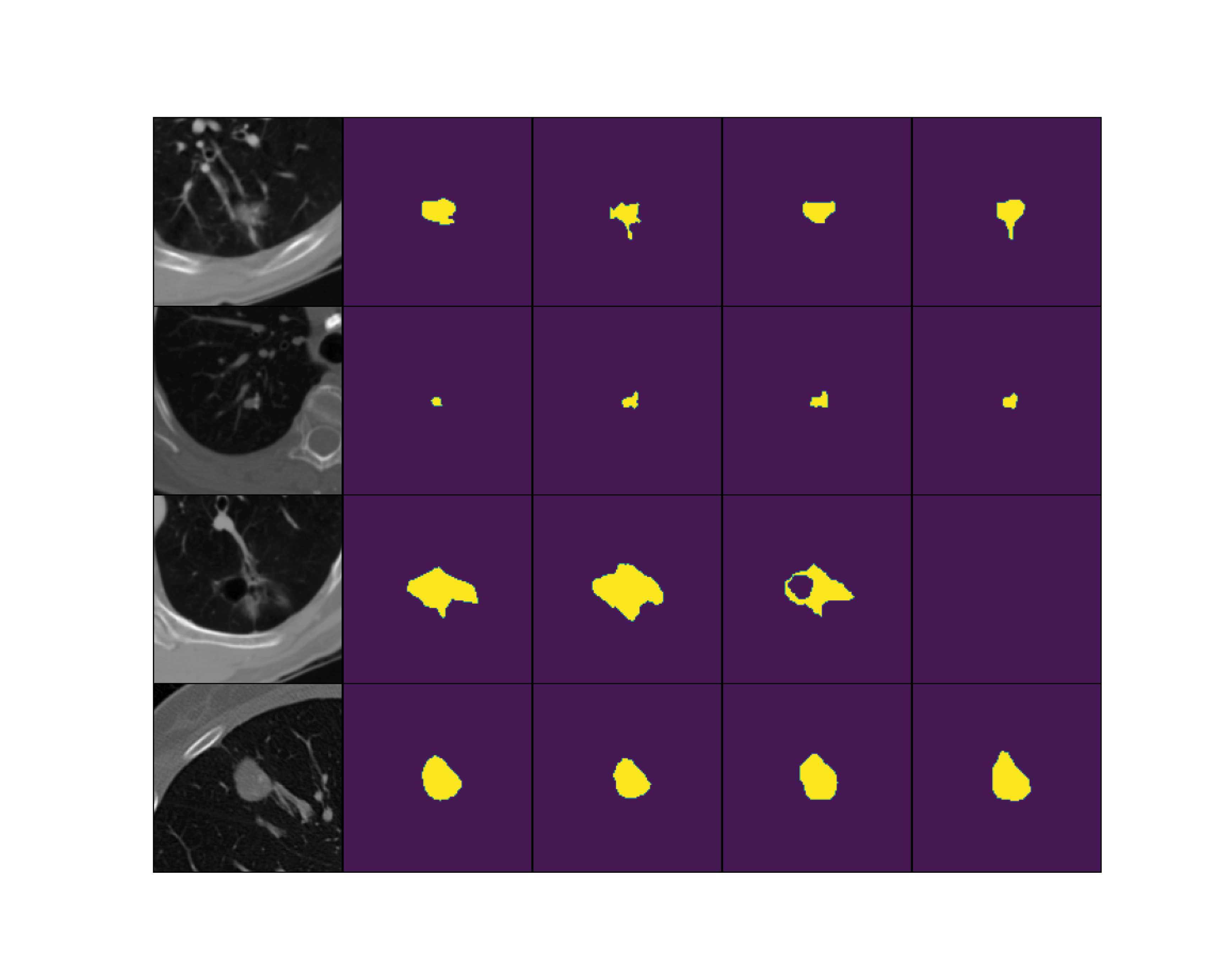}}
  \vspace{-0.9cm}
  \caption{\textit{Example images from the LIDC dataset.}}
  \label{example:LIDC}
\end{figure}

\begin{figure}
  \vspace{-0.5cm}
  \centerline{\includegraphics[width=0.7\linewidth]{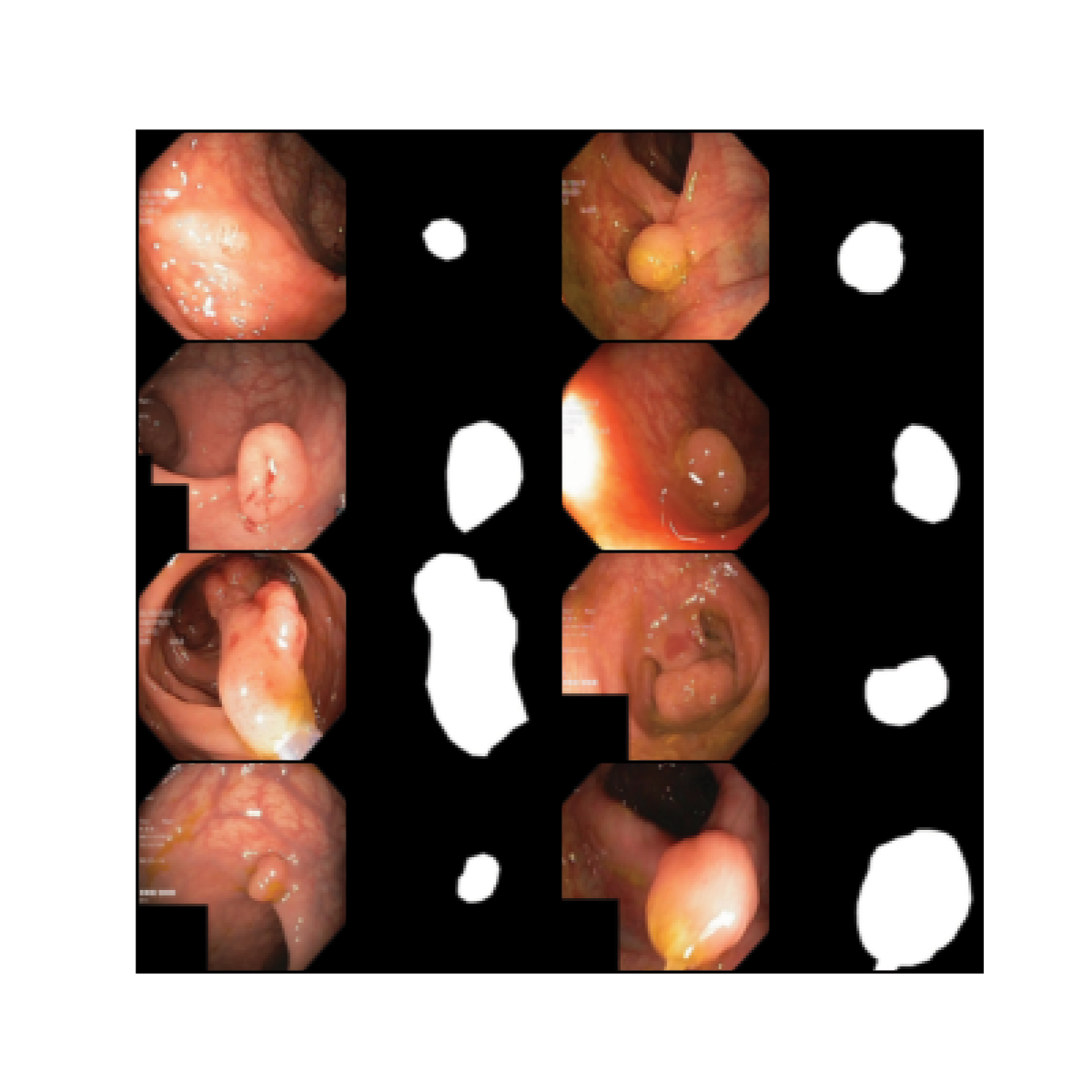}}
  \vspace{-1cm}
  \caption{\textit{Example images from the Kvasir-SEG dataset.}}
  \label{example:Kvasor}
\end{figure}

Here example images from the datasets used in this work can be seen. Figure \ref{example:LIDC} depicts four examples from the LIDC dataset. On the left in the figure the 2D CT image containing the lesion, followed by the four labels made by four independent annotators is shown. In Figure \ref{example:Kvasor}, eight examples from the Kvasir-SEG dataset is depicted. An endoscopic image with its ground truth label can be seen.

\newpage
\section{Sample size dependent GED}\label{gedgraph}
\begin{figure}
\centering
\begin{subfigure}{0.6\textwidth}
  \includegraphics[width=\linewidth]{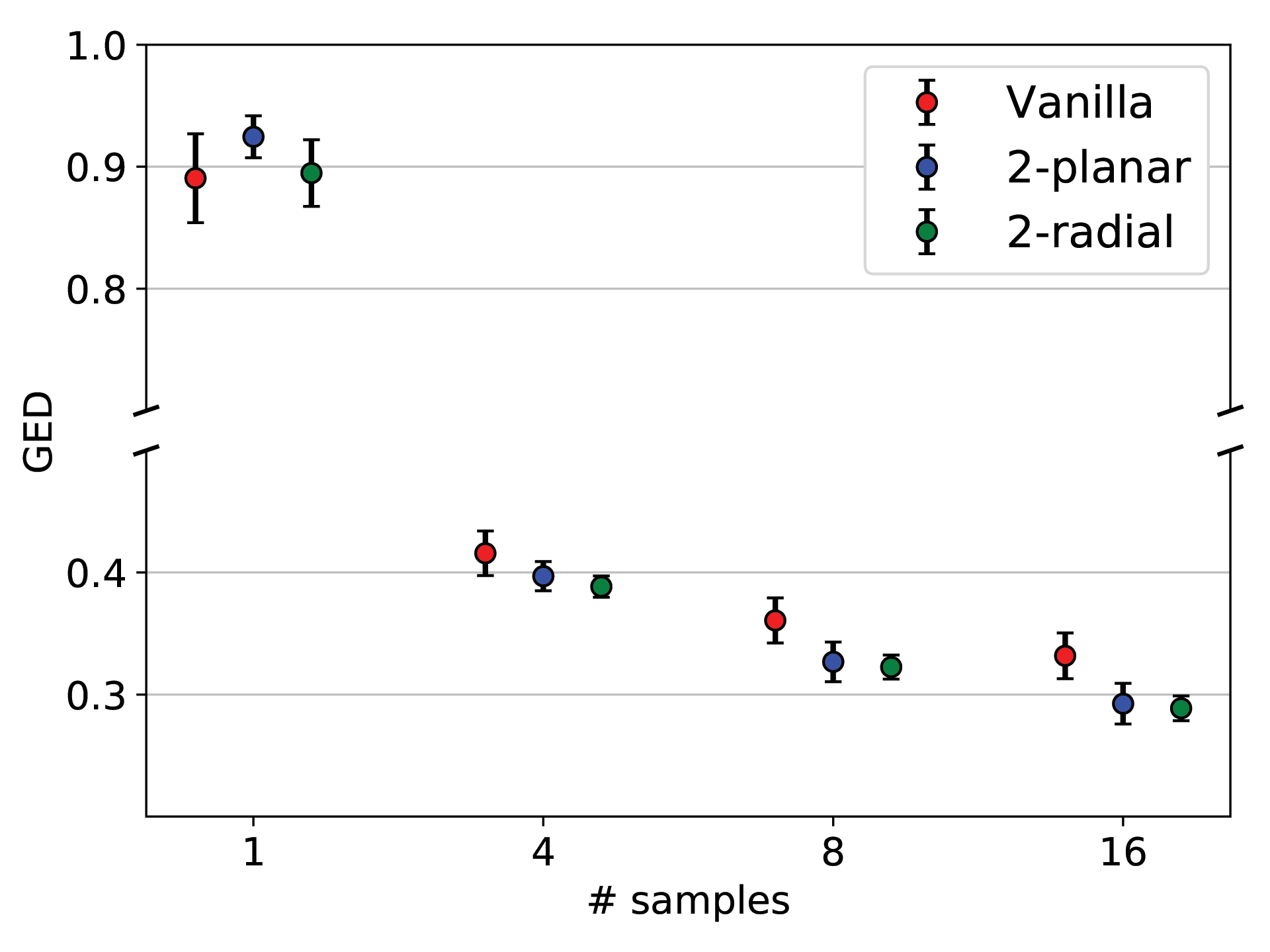}
  \caption{\textit{LIDC test set}}
  \label{fig:gedlidc}
\end{subfigure}%
\\
\begin{subfigure}{0.6\textwidth}
  \includegraphics[width=\linewidth]{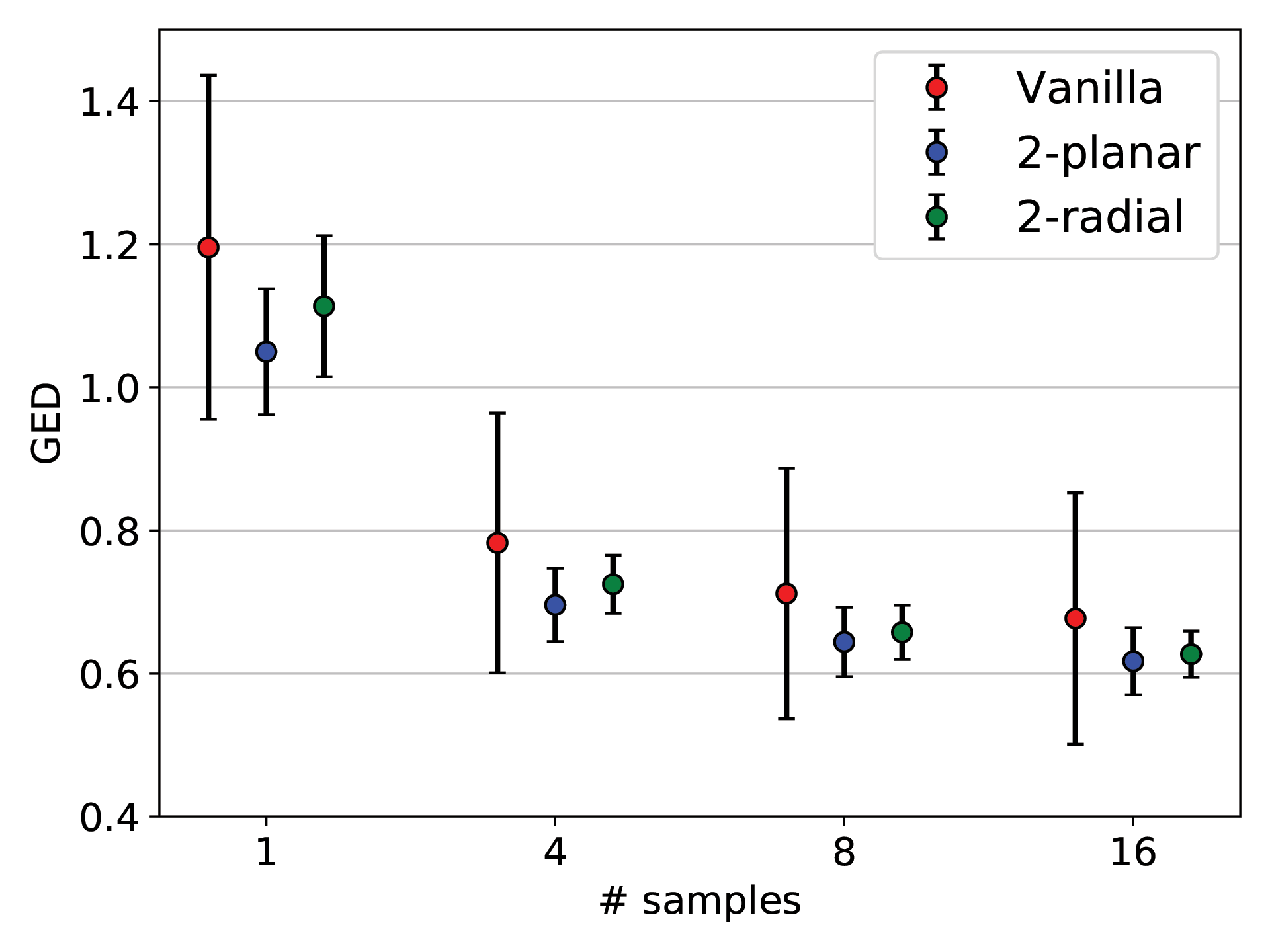}
  \caption{\textit{Kvasir-SEG test set}}
  \label{fig:gedkvasir}
\end{subfigure}
\caption{\textit{The GED based on sample size evaluated on the vanilla, 2-planar and 2-radial models.}}
\label{fig:ged}
\end{figure}

The GED evaluation is dependent on the number of reconstructions sampled from the prior distribution. Figure \ref{fig:ged} depicts this relationship for the vanilla, 2-planar and 2-radial posterior models. The uncertainty in the values originate from the changing results when training with ten-fold cross validation. One can observe that with increasing sample size, the GED as well as the associated uncertainty decrease. This is also the case when the posterior is augmented with a 2-planar or 2-radial flow. Particularly, the uncertainty in the GED evaluation significantly decreases.
\clearpage
\clearpage
\section{Prior distribution variance}\label{variance}
\begin{figure}
\begin{minipage}{\linewidth}
  \centering  \centerline{\includegraphics[width=\linewidth]{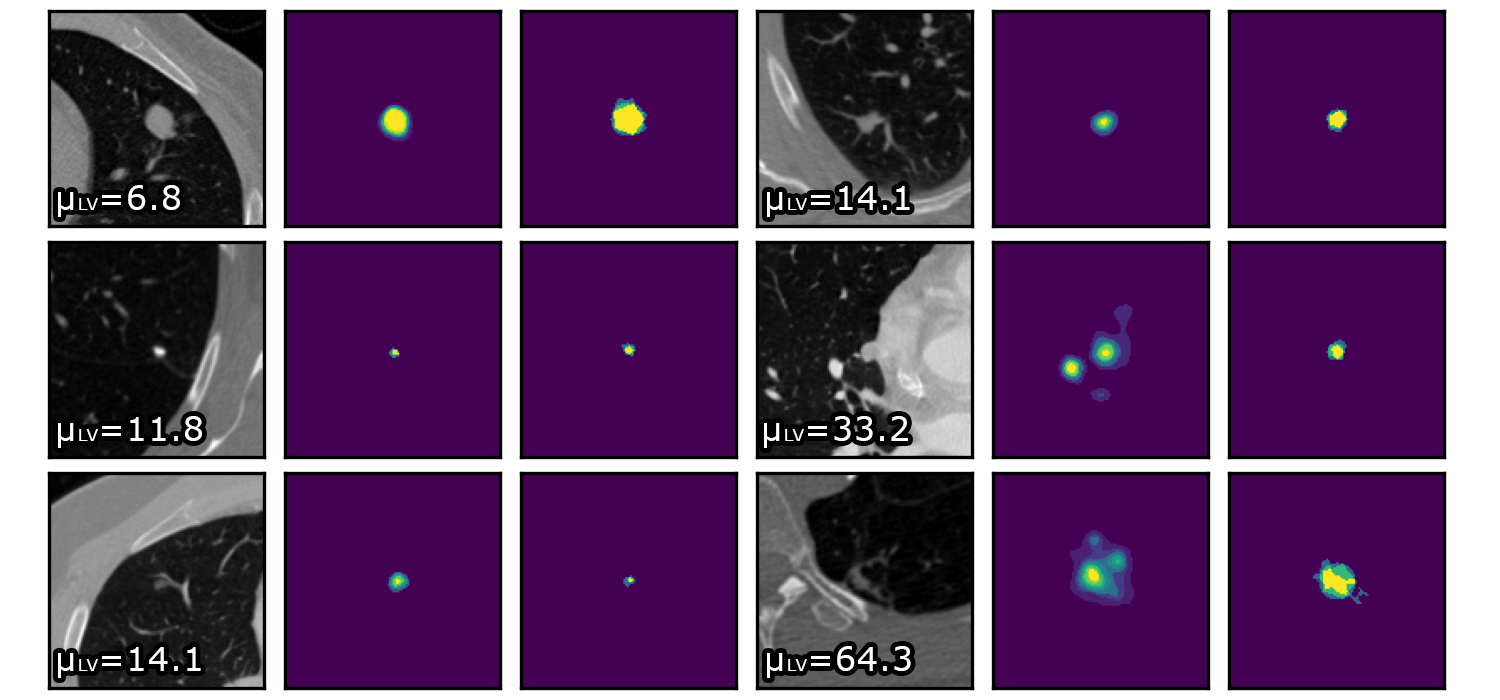}}
  \caption{\textit{Depicted in the CT image is the mean of the prior distribution variance of the 2-planar model. We show the input CT image, its average segmentation prediction (16 samples) and ground truth from four annotators.}}
  \label{fig:sigma}
\end{minipage}
\end{figure}

We investigated whether the prior distribution captures the degree of ambiguity in the input images. For every input image $\mathbb{X}$, we obtain a latent \textit{L}-dimensional mean and standard deviation vector of the prior distribution $P(\bm{\mu}, \bm{\sigma}\vert\mathbb{X})$.
The mean of the latent prior variance vector $\mu_{LV}$, is obtained from the input images in an attempt to quantify this uncertainty. Figure \ref{fig:sigma} shows this for several different input images of the test set. As can be seen, the mean variance over the latent prior increases along with a subjective assessment of the annotation difficulty.
\clearpage

\end{document}

%% file: introduction.tex
\section{Introduction}
\vspace{-0.25cm}
As a result of the considerable advances in machine learning research over the past decade, computer-aided diagnostics (CAD) using deep learning has rapidly been gaining attention. The outcome from these deep learning-based CAD systems has to be highly accurate, since it is often connected to critical designs resulting in a potentially large impact on patient care. As such, conclusions drawn from these CAD systems should be interpreted with care and by experts. A convolutional neural network (CNN)-based approach has been adopted in a large number of CAD applications and especially in semantic segmentation. This approach segments the objects of interest by assigning class probabilities to all pixels of the image. In the medical domain and especially in the context of lesion segmentation, the exact edges or borders of these lesions are not always easily defined and delineated by radiologists and endoscopists. In the case of multiple annotators, clinicians can also disagree on the boundaries of the localized lesions based on their understanding of the surrounding anatomy. However, the exact edges or borders of these areas of interests often play a critical role in the diagnostic process. For example, when determining whether to perform surgery on a patient or the surgical planning thereof, the invasion of a tumour into local anatomical structures derived from a CT scan, is crucial. Thus, multiple forms of uncertainties come into play with semantic segmentation-based approach for CAD. As such, accurately quantifying these uncertainties have become an essential addition in CAD. Specialized doctors provide ground-truth segmentation for models to be trained, based on their knowledge and experience. When this is done by multiple individuals per image, often discrepancies in the annotations arise, resulting in ambiguities in the ground-truth labels.

Recent work~\cite{KendallG17} suggests two types of uncertainties exist in deep neural networks. First, epistemic uncertainty, which refers to the lack of knowledge and can be minimized with information gain. In the case of multi/single-annotated data, these are the preferences, experiences, knowledge (or lack thereof) and other biases of the multiple/single annotators. This epistemic uncertainty from the annotator(s) manifests into aleatoric uncertainty when providing annotations. \textit{Aleatoric uncertainty} is the variability in the outcome of an experiment, due to the inherent ambiguity that exists in the data, captured through the multiple ground truths. By using a probabilistic segmentation model, we attempt to learn this as a distribution of possible annotations. It is important to enable expressiveness of the probability distributions to sufficiently capture the variability. In multi-annotator settings, the adoption of rich and multi-modal distributions may be more appropriate. We aim to show that by using invertible bijections, also known as Normalizing Flows (NFs), we can obtain more expressive distributions to adequately deal with the disagreement in the ground-truth information. Ultimately, this improves the ability to quantify the aleatoric uncertainty for segmentation problems.

In this work, we use the the Probabilistic U-Net (PU-Net) \cite{kohl2018probabilistic} as the base model and subsequently improve on it by adding a planar and radial flow to render a more expressive learned posterior distribution. For quantitative evaluation, we use the Generalized Energy Distance (GED), as is done in previous work (see related work). We hypothesize that this commonly used metric is prone to some biases, such that it rewards sample diversity rather than predictive accuracy. Therefore, we also evaluate on the average and Hungarian-matched IoU for the single- and multi-annotated data, respectively, as is also done by Kohl \textit{et al.}~\cite{kohl2019hierarchical}.


To qualitatively evaluate the ability to model the inter-variability of the annotations, we present the mean and standard deviation of the segmentation samples reconstructed from the model. In this paper, we make use of the multi- and single-annotated LIDC-IDRI (LIDC) and Kvasir-SEG datasets, thereby handling limited dataset size and giving insights on the effects of the complex posterior on hard-to-fit datasets. 

%% file: relatedwork.tex
\section{Related work}\label{related}
Kohl \textit{et al.}~\cite{kohl2018probabilistic} introduced the PU-Net for image segmentation, a model that combines the cVAE~\cite{sohn2015learning} and a U-Net~\cite{ronneberger2015u}. Here, the uncertainty is captured by a down-sampled axis-aligned Gaussian prior that is updated through the KL divergence of the posterior. These distributions contain several low-dimensional representations of the segmentation, which can be reconstructed by sampling. We use this model as our baseline model.

The concept of using NFs has been presented in earlier literature. For an extensive introduction to NFs we suggest the paper from Kobyzev~\textit{et al.}~\cite{Kobyzev_2020}. The planar and radial NFs have been used for approximating flexible and complex posterior distributions~\cite{rezende2016variational}. 

Selvan~\textit{et al.}~\cite{selvan2020uncertainty} used an NF on the posterior of a cVAE-like segmentation model and showed that this increases sample diversity. The increased sample diversity resulted in a better score on the GED metric and a slight decrease in DICE score. The authors reported significant gains in performance. However, we argue that this claim requires more evidence to confirm this positive effect, such as training with K-fold cross-validation and evaluating using other metrics. Also, insight  into the reasons for their improvements are not provided and critical details of the experiments are missing, such as the number of samples used for the GED evaluation. We aim to provide a more complete argumentation and show clear steps towards improving the quantification of aleatoric uncertainty.



%% file: architecture.tex
\section{Methods}\label{method}
\subsection{Model architecture}
\begin{figure}
\begin{minipage}{\linewidth}
  \centering  \centerline{\includegraphics[width=0.9\linewidth]{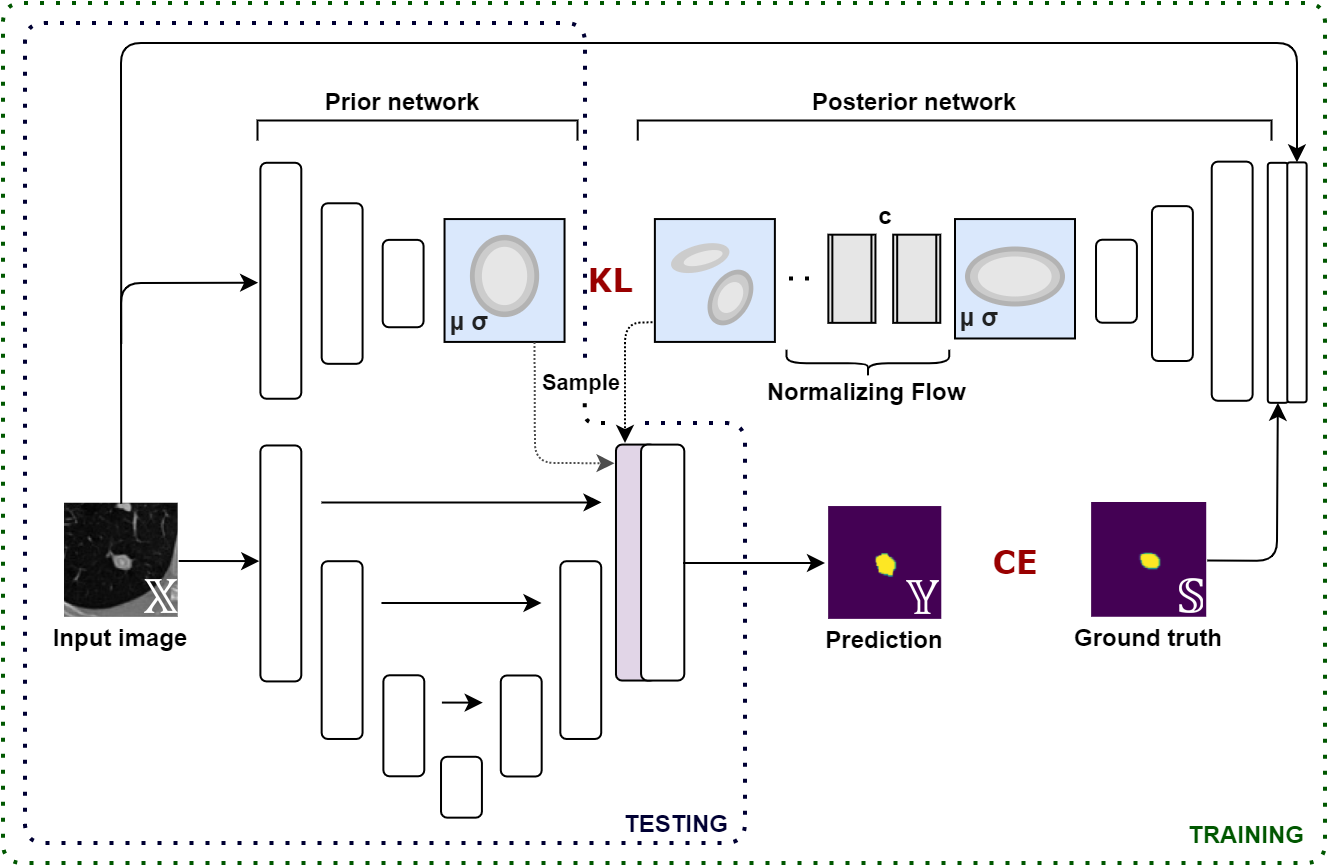}}
  \vspace{-0.2cm}
  \caption{\textit{Diagram of the PU-Net with a flow posterior.}}
  \vspace{-0.1cm}
  \label{fig:arch}
\end{minipage}
\end{figure}

\vspace{-0.2cm}
We use a PU-Net extended with an NF, as is shown in Figure~\ref{fig:arch}. A key element of the architecture is the posterior network $Q$, which attempts to encapsulate the distribution of possible segmentations, conditioned on the input image $\mathbb{X}$ and ground truth $\mathbb{S}$ in the base distribution. The flexibility of the posterior is enhanced through the use of an NF, which warps it into a more complex distribution.
During training, the decoder is sampled by the posterior and is constructing, based on the encoded input image, a segmentation via the proposed reconstruction network. The prior $P$ is updated with the evidence lower bound (ELBO \cite{kingma2014autoencoding}), which is based on two components: first, the KL divergence between the distributions $Q$ and $P$ and second, the reconstruction loss between the predicted and ground-truth segmentation. The use of NFs is motivated by the fact that a Gaussian distribution is too limited to fully model the input-conditional latent distribution of annotations. An NF can introduce complexity to $Q$, e.g. multi-modality, in order to more accurately describe the characteristics of this relationship.
\newpage\noindent
We proceed by extending the PU-Net objective (see appendix \ref{elboloss}) and explain the associated parameters in detail. We make use of the NF-likelihood objective (see Appendix~\ref{flows}) with transformation $f: \mathbb{R} \mapsto \mathbb{R}$ to define our posterior as
\begin{equation}
    \begin{aligned}
        \log q(\mathbf{z}\vert\mathbf{s},\mathbf{x})=\log q_{0}(\mathbf{z}_0\vert\mathbf{s},\mathbf{x})-\sum_{i=1}^{K} \log \left(\left|\operatorname{det} \frac{d f_{i}}{d \mathbf{z}_{i-1}}\right|\right).
    \end{aligned}
\end{equation}
to obtain the objective
\begin{equation}
    \begin{aligned}
        \mathcal{L} &= -\mathbb{E}_{q_\phi(\mathbf{z}_0\vert\mathbf{s},\mathbf{x})}[\,\operatorname{log}p(\mathbf{s}\vert\mathbf{z},\mathbf{x})\,] \\
        &+\operatorname{KL}\left(\,q_\phi(\mathbf{z}_0\vert\mathbf{s},\mathbf{x})\vert\vert p_\psi(\mathbf{z}\vert\mathbf{x})\,\right) -\mathbb{E}_{q_\phi(\mathbf{z}_0\vert\mathbf{s},\mathbf{x})}\left[\sum_{i=1}^{K} \log\left( \left|\operatorname{det} \frac{d f_{i}}{d \mathbf{z}_{i-1}}\right|\right)\right].
    \end{aligned}
    \label{lossfunc}
\end{equation}
The input-dependent context vector $\vect{c}$, is used to obtain the posterior flow parameters. During training, the posterior flow is used to capture the data distribution with the posterior network $Q(\bm{\mu}, \bm{\sigma}, \vect{c}\vert\mathbb{X},\mathbb{S})$, followed by sampling thereof to reconstruct the segmentation predictions $\mathbb{Y}$. At the same time, a prior network $P(\bm{\mu}, \bm{\sigma}\vert\mathbb{X})$ only conditioned on the input image is also trained through constraining its KL divergence with the posterior distribution. The first term in Eq.~(\ref{lossfunc}) entails the reconstruction loss, in our case the cross-entropy function as mentioned earlier. At test time, the prior network produces latent samples to construct the segmentation predictions.
\vspace{0.3cm}

%% file: method.tex
\vspace{-0.2cm}
\subsection{Data \& baseline experiments}
\vspace{-0.2cm}
We perform extensive experimental validation using the vanilla Probabilistic U-Net with 2-/4-step planar and radial flow variants on processed versions of the LIDC-IDRI (LIDC)~\cite{lidc} and the Kvasir-SEG~\cite{kvasir-seg} datasets. The preprocessed LIDC dataset~\cite{selvan2020uncertainty} transforms the 1,018 thoracic CT scans with four annotators into 15,096 128$\times$128-pixel patches according to the method of~\cite{kohl2018probabilistic,baumgartner2019phiseg}. Each image has 4 annotations. The Kvasir-SEG dataset contains 1,000 polyp images of the gastrointestinal tract from the original Kvasir dataset~\cite{kvasir}. We resize the images to be 128$\times$128 pixels as well and convert the images to gray-scale. Example images of the datasets can be found in Appendix~\ref{dataset}. As for the NFs, we used the planar flow conform related work~\cite{selvan2020uncertainty,berg2019sylvester} and also experiment with the radial flow. These flows are usually chosen because they are computationally the cheapest transformations that possess the ability to expand and contract the distributions along a direction (planar) or around a specific point (radial).
\vspace{-0.4cm}
\subsection{Performance evaluation}
\vspace{-0.2cm}
For evaluation, we deploy the \textit{Generalized Energy Distance} (GED) (also known as the \textit{Maximum Mean Discrepancy}), which is defined as
\begin{equation}
    \begin{aligned}
        D^2_{GED}(P_{pr}, P_{out})=2\mathbb{E}\left[d(\mathbb{S},\mathbb{Y})\right]-\mathbb{E}\left[d(\mathbb{S},\mathbb{S}')\right]-\mathbb{E}\left[d(\mathbb{Y},\mathbb{Y}')\right],
    \end{aligned}
\end{equation}
where $\mathbb{Y}$, $\mathbb{Y}'$ and  $\mathbb{S}$, $\mathbb{S}'$ are independent samples from the predicted distribution and ground truth distributions $P_{pr}$ and $P_{gt}$, respectively. Here, $d$ is a distance metric, in our case, one minus the Intersection over Union (1-IoU). When the predictions poorly match the ground truth, the GED is prone to simply reward diversity in samples instead of accurate predictions because the influence of the $\mathbb{E}\left[d(\mathbb{Y},\mathbb{Y}')\right]$ term becomes dominant. Therefore, we also evaluate the Hungarian-matched IoU, using the average IoU of all matched pairs for the LIDC dataset. We duplicate the ground-truth set, hence matching it with the sample size. Since the Kvasir-SEG dataset only has a single annotation per sample, we simply take the average IoU from all samples. Furthermore, when the model correctly predicts the absence of a lesion (i.e. no segmentation), the denominator of the metric is zero and thus the IoU becomes undefined. In previous work, the mean excluding undefined elements was taken over all the samples. However, since this is a correct prediction, we award this with full score (IoU=~1) and compare this approach with the method of excluding undefined elements for the GED.

To qualitatively depict the model performance, we calculate the mean and standard deviation with Monte-Carlo simulations (i.e. sampling reconstructions from the prior). All evaluations in this paper are based on 16~samples to strike a right balance between sufficient samples and a justifiable approximation, while maintaining minimal computational time.
\vspace{-0.4cm}
\subsection{Training details}
\vspace{-0.2cm}
The training procedure entails tenfold cross-validation using a learning rate of $10^{-4}$ with early stopping on the validation loss based on a patience of 20 epochs. The batch size is chosen to be~96 and~32 for the LIDC and the Kvasir-SEG dataset, respectively. The dimensionality of the latent space is set to $L=6$. We split the dataset as 90-10 (train/validation and test) and evaluate the test set on the proposed metrics. All experiments are done on an 11-GB RTX 2080TI GPU.

%% file: results.tex
\section{Results and discussion}\label{results}
\subsection{Quantitative evaluation}
\vspace{-0.2cm}
We refer to the models by their posterior, either unaugmented (vanilla) or with their \textit{n}-step Normalizing Flow (NF). The results of our experiments are presented in Table~\ref{tab:results}. In line with literature, it shown that the GED improves with the addition of an NF. This hypothesis is tested using both a planar and radial NF and observe that both have a similar effect. Furthermore, both the average and Hungarian-matched IoU improve with the NF. We find that the 2-step radial (2-radial) NF is slightly better than other models for the LIDC dataset, while for the Kvasir-SEG dataset the planar models tend to perform better. The original PU-Net introduced the variability capturing of annotations into a Gaussian model. However, this distribution is not expressive enough to efficiently capture this variability. The increase in GED and average IoU performance from our experiments confirm our hypothesis that applying NF to the posterior distribution of the PU-Net improves the accuracy of the probabilistic segmentation. This improvement occurs because the posterior becomes more complex and can thus provide more meaningful updates to our prior distribution.

Including/excluding correct empty predictions did not result in a significant difference in the metric value when comparing the vanilla models with the posterior NF models. Our results show that the choice in NF has minimal impact on the performance and suggest practitioners to experiment with both NFs. Another publication in literature~\cite{selvan2020uncertainty} has experimented with more complex posteriors such as GLOW~\cite{kingma2018glow}, where no increase in performance was obtained. In our research, we have found that even a 4-step planar or radial NF (which are much simpler in nature) can already be too complex for our datasets. 
A possible explanation is that the variance in annotations captured in the posterior distribution only requires a complexity that manifests from two NF steps. This degree of complexity is then most efficient for the updates of the prior distribution. More NF steps would then possibly introduce unnecessary model complexity as well parameters for training, thereby reducing the efficiency of the updates. Another explanation could be that an increase in complexity of the posterior distribution does in fact model the annotation variability in a better way. Nevertheless, not all information can be captured by the prior, as it is still a Gaussian. In this case, a 2-step posterior is close enough to a Gaussian for meaningful updates yet complex enough to be preferred over a Gaussian distribution. We consider that for similar problems, it is better to adopt simple NFs with a few steps only. However, we can imagine the need for a more complex NF for other scenarios, where the varying nature in the ground truth follows different characteristics, e.g. encompassing non-linearities.

\begin{table}
\centering
{\resizebox{0.85\linewidth}{!}{
\begin{tabular}{cccccc} 
\toprule
\multirow{2}{*}{\underline{Dataset} \enspace}    & \multirow{2}{*}{\underline{Posterior} \enspace} & \multicolumn{2}{c}{\textbf{GED}} & \multicolumn{2}{c}{\textbf{IoU}}  \\ 

                                &                           & \textit{Excl.}    & \textit{Incl.}        & \textit{Avg.}     & \textit{Hungarian}     \\ 
\hline
\multirow{3}{*}{LIDC \enspace}      & Vanilla  \enspace  & 0.33 $\pm$ 0.02 \enspace & 0.39 $\pm$ 0.02 \enspace              & --- \enspace               & 0.57 $\pm$ 0.02 \\
                                    & 2-planar \enspace  & 0.29 $\pm$ 0.02 \enspace & 0.35 $\pm$ 0.03 \enspace     & --- \enspace  & 0.57 $\pm$ 0.01 \\
                                    & 2-radial \enspace  & \textbf{0.29 $\pm$ 0.01} \enspace & \textbf{0.34 $\pm$ 0.01} \enspace     & --- \enspace  & \textbf{0.58 $\pm$ 0.01} \\
                                    & 4-planar \enspace  & 0.30 $\pm$ 0.02 \enspace        & 0.35 $\pm$ 0.04 \enspace              & --- \enspace               & 0.57 $\pm$ 0.02 \\
                                    & 4-radial \enspace  & 0.29 $\pm$ 0.02 \enspace & 0.34 $\pm$ 0.03 \enspace     & --- \enspace  & 0.57 $\pm$ 0.01 \\
\hline
\multirow{3}{*}{Kvasir-SEG\enspace}    & Vanilla  \enspace  & 0.68 $\pm$ 0.18 \enspace             & 0.69 $\pm$ 0.17 \enspace              & 0.62 $\pm$ 0.07 \enspace  & --- \\
                                    & 2-planar \enspace  & 0.62 $\pm$ 0.05 \enspace    & 0.63 $\pm$ 0.05 \enspace     & \textbf{0.71 $\pm$ 0.01} \enspace   & --- \\
                                    & 2-radial \enspace  & \textbf{0.63 $\pm$ 0.03} \enspace    & \textbf{0.64 $\pm$ 0.03} \enspace     & 0.66 $\pm$ 0.06 \enspace   & --- \\
                                    & 4-planar \enspace  & 0.63 $\pm$ 0.06 \enspace             & 0.67 $\pm$ 0.05 \enspace              & 0.71 $\pm$ 0.04 \enspace  & --- \\
                                    & 4-radial \enspace  & 0.65 $\pm$ 0.04 \enspace    & 0.67 $\pm$ 0.05 \enspace     & 0.65 $\pm$ 0.07 \enspace   & --- \\
\bottomrule
\end{tabular}}
    \vspace{-0.2cm}
    \caption{\textit{Test set evaluations on the GED and IoU based on 16 samples. Further distinction in the GED is made on whether the correct empty predictions are included. The IoU is evaluated with the Hungarian-matching algorithm and averaged with the LIDC and Kvasir-SEG dataset, respectively.\label{tab:results}}}}
    \vspace{-1cm}
\end{table}
We have also compared the vanilla, 2-planar and 2-radial models by depicting their GEDs based on sample size (Appendix~\ref{gedgraph}). As expected, the GEDs decrease as the number of samples increase. It is also evident that the variability in metric evaluations is less for models with an NF posterior. The NF posteriors consistently outperform the vanilla PU-Net for the LIDC and Kvasir-SEG datasets. 
\subsection{Qualitative evaluation}
\begin{figure}[h!]
\centering
\begin{minipage}{0.9\linewidth}
  \centerline{\includegraphics[width=\linewidth]{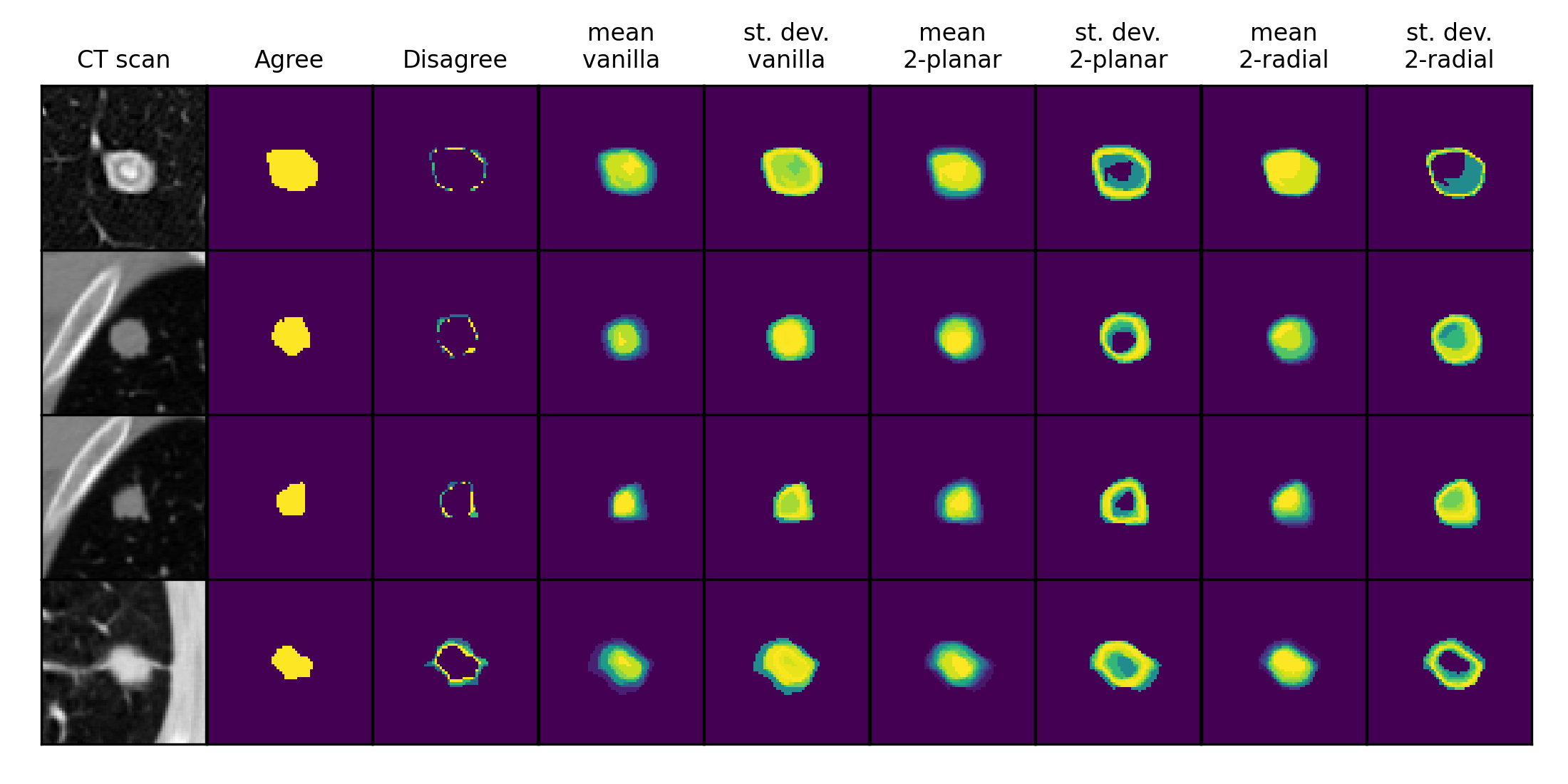}}
  \vspace{-0.5cm}
  \caption{\textit{Reconstructions of the LIDC test set.}}
  \label{fig:lidcsample}
  \centerline{\includegraphics[width=\linewidth]{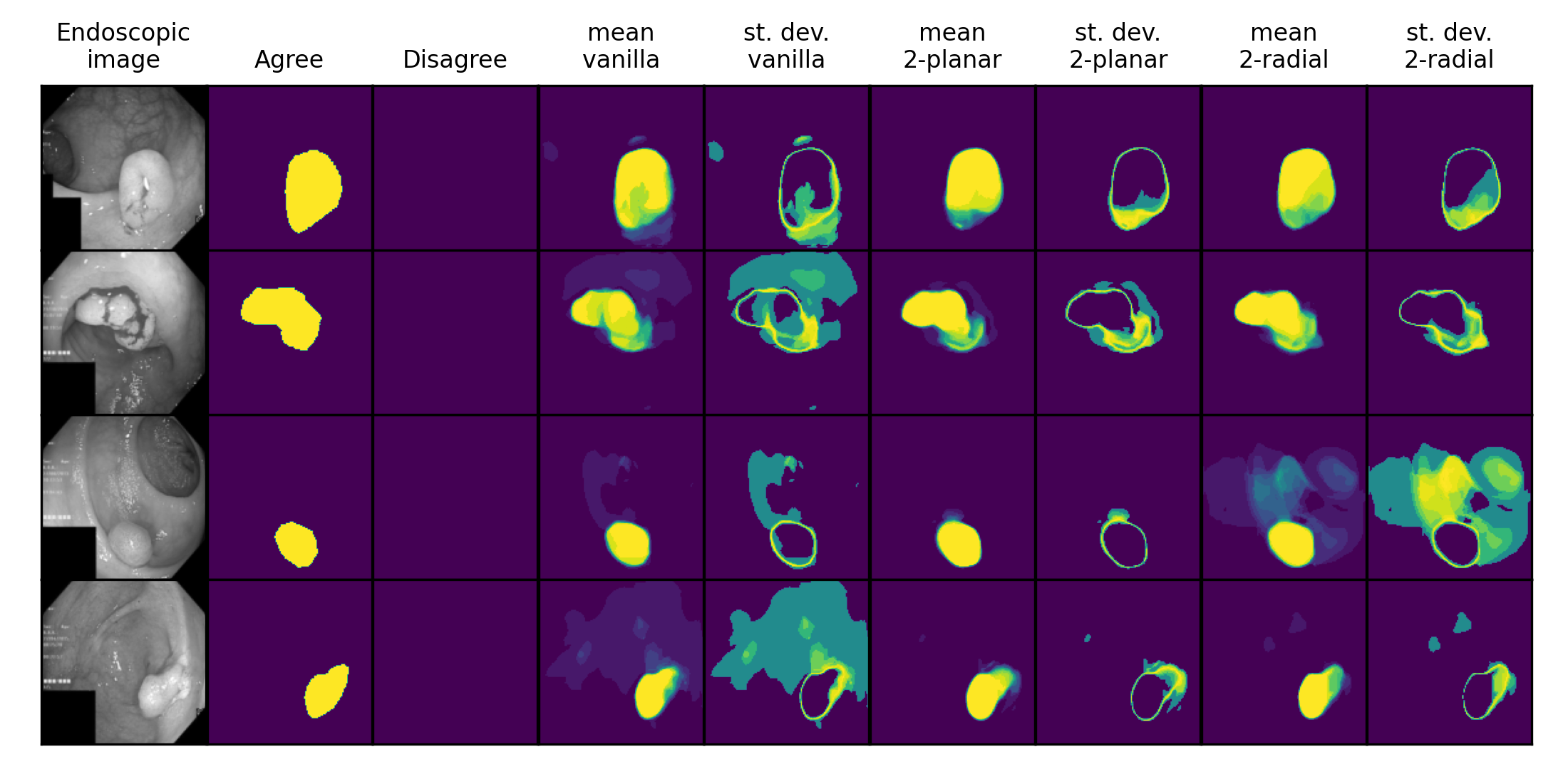}}
  \vspace{-0.5cm}
  \caption{\textit{Reconstructions of the Kvasir-SEG test set.}}
  \label{fig:kvasirsample}
\end{minipage}
\vspace{-0.4cm}
\end{figure}

\vspace{-0.2cm}
The mean and standard deviation based on 16~segmentation reconstructions from the validation set is shown in Figures~\ref{fig:lidcsample} and~\ref{fig:kvasirsample}. Ideally, it is expected to obtain minimal uncertainty at the center of the segmentation, because annotations agree on the center area most of the time for our datasets. This also implies that the mean of the center should be high because of the agreement of the annotators. For both datasets, the means of our sampled segmentations match well with the ground truths and have high values in the center areas corresponding to good predictions. Furthermore, the PU-Net without an NF shows uncertainty at both edges and segmentation centers. In contrast, for all NF posterior PU-Net models, the uncertainty is mostly on the edges alone. A high uncertainty around the edges is also expected, since at those areas the annotators almost always disagree. From this, we can conclude that NF posterior models are better at quantifying the aleatoric uncertainty of the data. Even though there is no significant quantitative performance difference between the NF models, there is a well distinguishable difference in the visual analysis. In almost all cases, it can be observed that the planar is better than the radial NF posterior in learning the agreement between the segmentation centers. We also investigated the prior distribution to determine if it captures the ambiguity that exist in the input image. In Appendix~\ref{variance}, we show the prior distribution variance for different test set input images. We qualitatively observe that with increasing variance, the subjective assessment of the annotation difficulty increases. This suggests the possibility of obtaining an indication of the uncertainty in a test input image without sampling and evaluating the segmentation reconstructions. 

The prior distribution is an area that needs to be further explored, since this is still assumed  Gaussian. We hypothesize that augmenting the prior with an NF could result in further improvements. Future work will also include an investigation into the correlation between the prior and segmentation variance. A limiting factor of the proposed model is the use of only a single distribution. We consider that when using flexible distributions at multiple scales, the overall model will further improve.

%% file: conclusion.tex
\section{Conclusion}
\vspace{-0.25cm}
Quantifying uncertainty in image segmentation is very important for decision-making in the medical domain. In this paper, we propose to use the broader concept of Normalizing Flows (NFs) for modeling both single- and multi-annotation data. This concept allows more complex modeling of aleatoric uncertainty. We consider modeling of the posterior distribution by Gaussians too restrictive to model variability. By augmenting the model posterior with a planar or radial NF, we attain up to 14\% improvement in GED and 13\% in IoU, resulting in a better quantification of the aleatoric uncertainty. We propose that density modeling with NFs is something that should be experimented with throughout other ambiguous settings in the medical domain, since we are confident this will result in valuable information for further research. A significant improvement has been found through only augmenting the posterior distribution with NFs, whereas little-to-none investigations have been made into the effect of additionally augmenting the prior distribution and is suggested for future work. Moreover, we suggest augmenting other architectures that aim to capture uncertainty and variability through a learnt probability distribution with Normalizing Flows.